# Detection of Racial Bias from Physiological Responses

Fateme Nikseresht [1], Runze Yan [1], Rachel Lew [1], Yingzheng Liu [1], Rose M.Sebastian [1]
and Afsaneh Doryab [1]

[1]University of Virginia, USA
{fn5an, ry4jr, rel7dqj, yl4dt, rs9rf, ad4ks}@virginia.edu

**Abstract.** Despite the evolution of norms and regulations to mitigate the harm from biases, harmful discrimination linked to an individual's unconscious biases persists. Our goal is to better understand and detect the physiological and behavioral indicators of implicit biases. This paper investigates whether we can reliably detect racial bias from physiological responses, including heart rate, conductive skin response, skin temperature, and micro-body movements. We analyzed data from 46 subjects whose physiological data was collected with Empatica E4 wristband while taking an Implicit Association Test (IAT). Our machine learning and statistical analysis show that implicit bias can be predicted from physiological signals with 76.1% accuracy. Our results also show that the EDA signal associated with skin response has the strongest correlation with racial bias and that there are significant differences between the values of EDA features for biased and unbiased participants.

## 1    Introduction

Unconscious or implicit biases, which are unfair "prejudice[s] in favor of or against one thing, person, or group compared with another" [14], continue to be a pressing social issue. People can hold biases against or for others based on characteristics such as the individual's age, gender, gender identity, physical abilities, religion, sexual orientation, or weight [14]. Due to their automatic and unintentional nature – and often unrecognized impact on judgment and behavior – implicit biases can lead people, even those most committed to egalitarian ideals, to be unsuspecting perpetrators of discrimination.

Existing bias interventions have two primary shortcomings: (1) they have been designed as "one-shot", "one size fits all" solutions and (2) they have not successfully aided individuals in identifying bias-triggering contexts or situations as they occur in real time. These shortcomings are in large part due to the difficulty and lack of feasibility of assessing or detecting bias as it manifests itself in daily life.

Our research aims to understand and detect the physiological and behavioral associates of implicit bias and, ultimately, to build technology that uses this knowledge to integrate bias mitigation strategies into personal devices, such as smartphones and wearables. In this study, we use machine learning modeling to understand the links between racial bias and involuntary physiological responses using Photoplethysmography (PPG), Electrodermal Activity (EDA), Skin

Temperature (SKT), and Accelerometer (ACC) signals. To our knowledge, this is the first study to investigate the feasibility of using machine learning to detect racial bias from physiological signals collected from a wearable wristband.

In the following sections, we discuss prior research on implicit bias and its physiological counterparts. We then describe our data collection study, our analysis methods, and our preliminary results.

## 2    Background and Related Work

### 2.1    Detecting bias via the Implicit Association Test (IAT)

Implicit biases are typically measured through the Implicit Association Test (IAT) [2], a test that measures how long it takes participants to categorize contrasted concepts, like Black and good, or white and kind [11]. For example, a participant is described as having an unconscious bias towards white appearing faces if they are faster at completing the task when white faces are paired with words like "good" and Black appearing faces are paired with words like "bad." Based on the differences in participants' speed in pairing faces and evaluations in Black-good/White-bad and Black-bad/White-good conditions, participants are categorized as having no bias, a slight bias, a moderate bias, or a strong bias [12]. A meta-analysis of 122 studies of the IAT found that the predictive validity of the IAT is greater for social sensitive topics, such as race, where the test has greater predictive validity for bias than self-report measures [11]. Even small differences between individuals on the IAT have been linked to differences in discrimination in hiring, in salary, in criminal proceedings, in school discipline and grading, and in health care decisions [10]. What the IAT fails to tell us, however, is whether there are also physiological responses associated with bias.

### 2.2    Physiological Sensing and its Relation to Bias

Our ultimate goal is to identify the triggers and internal indicators of bias so that we can provide individuals with just-in-time information they can use to reduce the impact of biases on their judgments and actions. Prior work has begun to establish one avenue for such interventions: the linking of unconscious bias activation with distinct patterns of concomitant physiological activity, patterns that are consistent with threat responses in intergroup contexts. Researchers have found that learned threat responses to outgroup members are indicated by distinct patterns of neural activity (e.g., amygdala activation: [3, 4]); cardiovascular reactivity [6, 13]; and skin conductance [8]). For example, the intensity of participants' neural activity is related to differences in assessed levels of prejudice. Participants' anterior cingulate cortex activity is greater for participants with more assessed prejudice than their less prejudiced peers when they are asked to engage in a labelling activity after being shown Black appearing faces [5]. Similarly, in another study, when participants were confronted with individuals who appeared Asian, but had Southern American accents, the participants demonstrated unusual cardiovascular responses, had poorer task performance, and manifested negative and defeat-related behaviors [6]. In studies of bias against individuals with mental illness, researchers found that the reactions and decisions of participants when interacting with individuals with mental health concerns were correlated with skin conductance measures, indicating a possible relationship between skin conductance measures and bias [8].

While these early studies on the relationship between implicit biases and physiological responses are encouraging, our research is motivated by gaps in the literature. We are not aware of any research in this domain that captures and investigates a combination of physiological responses and measures of implicit biases. Our study helps to 1) better understand and confirm that such relationships exist, 2) identify the most significant physiological responses to bias, and 3) identify the overlapping and combined physiological responses to bias.

## 3 Methods

### 3.1 Data Collection

We used Empatica [1], a wristband designed to gather high-quality physiological signals to capture physiological responses. The Empatica has four sensors: Electrodermal Activity (EDA), Photoplethysmography (PPG), Infrared Thermopile (TEMP), and Accelerometer (ACC). The PPG sensor is capable of measuring Blood Volume Pulse (BVP) from which Heart Rate (HR), Heart Rate Variability (HRV), Inter-Beat-Interval (IBI) and many other cardiovascular features can be extracted. The EDA sensor is used to measure sympathetic nervous system arousal and monitor emotional states, such as stress, excitement, and focus. The accelerometer and Infrared Thermopile are used to track body movements and to monitor changes in skin temperature.

As part of a larger study on simulated practice, the data collection took place in two of eight sections of an Introduction to Teaching course at a large university in the Southeast. Of the 76 students in the two sections, 46 consented to wearing an Empatica E4 wristband while completing the study measures and had enough biometric data for analysis. All participants completed the IAT and surveys during course sections, observed by the researchers.

The participants resembled their course peers in gender, interest in teaching, and languages spoken at home, but were slightly more white than the students in the course sections as a whole (61% to 51%) and from slightly wealthier families (54% to 42%). Overall, the participants were almost equally split between men and women, generally spoke only English in the home, and were more likely to have attended a primarily white high school than a mixed race school or one primarily consisting of students of color (50% primarily white, 39% mixed race, and 11% primarily students of color).

### 3.2 Data Processing

**IAT data processing.** The initial IAT test results had eight categories. The IAT categorized participants as having either a strong preference, moderate preference, slight preference, or no preference for either white or Black appearing faces. We designed our machine learning analysis as a binary classification task to infer whether or not a person has bias towards any racial group. We, therefore, categorized all strong and moderate preferences as 'biased' and all slight or no preferences as 'unbiased'. Out of the 46 participants, 26 scored as biased and 20 scored as unbiased.

**Feature Extraction.** We calculated a total of 91 features from SKT, HR, BVP, EDA, and ACC signals as described in Table 1. Statistical features such as *maximum, mean*, and *standard deviation* were common across all signals, and other features such as *rms* and *skewness* were extracted from EDA and BVP signals.

The temperature and blood volume pulse signals (BVP) from the Empatica were measured at a sampling frequency of 4 Hz and 64 Hz respectively. Heart rate (HR) was derived from BVP signals by an algorithm built into the Empatica E4. The 3-axis acceleration signal was measured at a sampling frequency of 32 Hz. Because we were more interested in the degree of movement than the direction of the movement, we derived the magnitude of the *x, y,* and *z* axes from the accelerometer signals using the following equation:

$$Magnitude = \sqrt{x^2 + y^2 + z^2}$$

The EDA signals were measured at a sampling frequency of 4 Hz. The cvxEDA library [9] was used to separate the phasic and tonic components of EDA signals. This allowed us to extract features from the main, phasic, and tonic parts of the EDA signal, which is helpful in analyzing stimulus responses. The tonic part of the signal is more correlated to physiological arousal and alertness, while the phasic component is more connected to attention, significance, and novelty [7]. The features of EDA and BVP signals, including *rms, kurtosis, skewness, zero_cross*, and *pow_spec* were extracted using Python libraries HeartPy [15] and pyphysio [5].

### 3.3 Machine Learning Analysis

We used machine learning to understand 1) how accurately we could classify the biased group from the unbiased sample, 2) what physiological responses were most indicative of the existence of racial bias in individuals, and 3) what temporal patterns could be observed in the physiology of the sample population during the IAT that might provide useful insights into the onset of an individual's bias-reaction.

We used XGBoost, a machine learning ensemble meta-algorithm that uses decision trees as base-learner algorithms, under the framework of Bootstrap Aggregation (Bagging) and Boosting mechanisms. This method can also report the ranking of each feature from its average importance and its contribution to the algorithm's decision-making process. Since the number of biased and unbiased participants was imbalanced (26 biased vs. 20 unbiased), we applied oversampling technology in the leave-one-participant-out cross validation to balance the number of participants in both groups. We divided the signal stream of each sensor into 5 second intervals and extracted features as discussed in previous sections. We used this short time window to capture fine-grained and micro features from data. This process generated a dataset of 5-second samples as rows and sensor features as columns. We then applied our machine learning algorithms on the 5-second sample dataset in a leave-one-person-out cross validation. This approach classified each 5-second sample of each participant's data into the two labels of 0 (unbiased) or 1 (biased). The process resulted in sequences of 0 and 1s for each participant (see Figure 1). We developed an algorithm to iteratively smooth the sequence and generate consecutive blocks of 0 and 1 labels. First, the sequence was parsed and blocks of labels with length of 1 were replaced with the neighboring majority label on both sides. For example, 1101 was smoothed to 1111. Then segments with length of 2 were checked and smoothed in the same way. The process continued until either at most three consecutive windows were created or the frequencies of the remaining segments were greater than the mean value of the original segments' frequencies. We then chose the label that had the largest consecutive block in the sequence as the final label.

**Table 1.** Extracted features from physiological signals (EDA, BVP, HR, SKT, and Magnitude (extracted from Accelerometer)).

| Signal | Feature | Description |
|---|---|---|
| EDA, EDA tonic, EDA phasic, BVP, HR, SKT, Magnitude | Statistical Features | max, min, median, mean, std, var, interq_range |
| | mean_abs_dev | Mean absolute deviation of the signal. |
| | distance | Total distance traveled by the signal using the hipotenusa between two data points. |
| EDA, EDA tonic, EDA phasic, BVP | rms | Root means square of the signal. |
| | *kurtosis* | Measures the peakedness of the probability density function (PDF) of a time series. |
| | skewness | Indicates the symmetry of the PDF of the amplitude of a time series. |
| | zero_cross | Corresponds to the total number of times that the signal changes from positive to negative or vice versa. |
| | power_spec | Maximum power spectrum density of the signal. |
| | num_peaks | Number of peaks. |
| | auc | The area under the curve of the signal computed with trapezoid rule. |
| EDA, EDA tonic, EDA phasic | max_peak | Maximum peak values. |
| BVP | mean_peak | Mean of peak values. |
| | sdnn | Standard deviation of NN intervals. |
| | sdsd | Standard deviation of successive RR intervals differences. |
| | rmssd | Root mean square of successive RR intervals differences. |
| | pnn20 /pnn50 | Percentage of successive RR intervals that differ by more than 20/50 ms. |
| | hr_*mad* | Median absolute deviation of RR intervals. |
| | sd1/sd2 | Ratio of *sd*1 to *sd*2. |
| | *breathingrate* | Number of breaths per minute. |

This strategy generated the ranking of features during the cross validation. We further analyzed the set of features that had a high importance rate in more than half of the iterations. To measure the performance of the algorithms, we calculated accuracy, F1 score, precision, and recall statistics for the biased and unbiased groups.

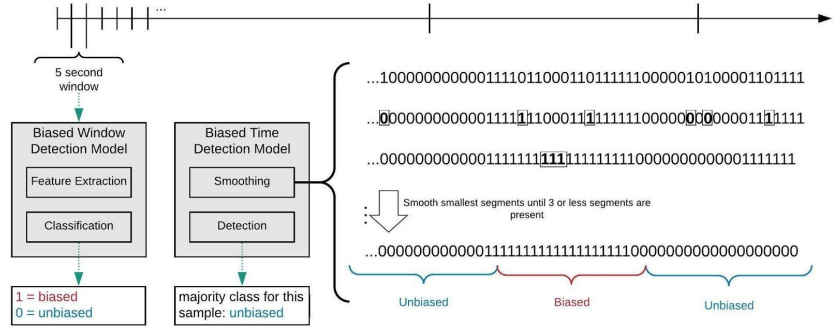

**Fig. 1.** Smoothing process

## 4 Results

Our machine learning analysis focused on 1) demonstrating the feasibility of using physiological responses to measure racial bias, 2) identifying physiological features and their importance in influencing the decision of algorithms in classifying biased vs. unbiased participants, and 3) identifying patterns of significant change in participants' time series data that might relate to their racial bias.

In addition to the accuracy of classifying biased vs. unbiased samples, we wanted to identify temporal patterns that might emerge in participants' physiology during the IAT. With the ratio of majority class (biased samples) as baseline (56.5%), our results indicated that XGBoost provided an overall accuracy of 19.6% above the baseline respectively (Accuracy = 76.1% and F1 =75.8%,). The algorithms could accurately label the majority of samples belonging to the biased class (Recall of 76.9%) compared to the unbiased class, which was expected given that the majority of the samples belong to the biased class. The higher impact of accurate classification of samples in the biased class to some degree compensates for more false negatives in the unbiased samples (Recall of 75%).

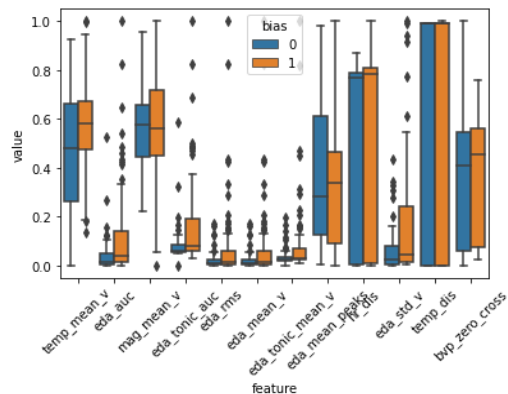

**Fig. 2.** Frequently highly ranked features based on XGBoost algorithm.

Figure 2 shows the features that had highest ratings in the majority of the iterations during the cross-validation. We only show the features with the frequency above 23, which is half the number of participants (46). The majority of those features were from the EDA signals.

To understand the temporal patterns in the physiological responses, we looked at the location of the majority window of true predicted labels in smoothed sequences belonging to participants whose sequence was accurately labeled as biased to see if any patterns emerged. For 66% of participants, the majority window was placed at the end of the sequence, indicating that physiological reactions in biased participants became stronger as the test progressed. While more analyses will be necessary to draw conclusions on this observation, this finding suggests that timing and duration of physiological responses could serve as additional indicators of bias reactions.

## 5   Discussion

Our machine learning analyses provide preliminary evidence of the feasibility of using physiological responses to measure and understand racial bias. The EDA measures of skin conductivity had the strongest related to bias. The EDA measures constituted the majority of highly ranked features, and the EDA measure of standard deviation differed the most between the biased and unbiased groups. Further, the increase in participants' physiological reactions over the duration of the IAT indicates that bias responses themselves might increase over prolonged interactions, which has important practical implications for designing bias interventions. Our findings on skin conductivity demonstrate the need for more research on physiological indicators and bias, so that we can learn if this finding is limited to just the race IAT, just these participants, or if the finding is broadly true for more biases, situations, and samples.

While the findings of this study have broad implications for the study of implicit bias, they are only preliminary and more research is needed to confirm them. The sample for this study consists of only 46 students, and the findings might vary with a larger sample. In addition, the small sample size created a severe class imbalance and limited our ability to do multi-class classification to infer different levels of bias. While we mitigated this problem by grouping the data samples into two categories of biased and unbiased, ideally, we would only use data samples that were labeled as highly biased or not biased to have a clear distinction between the two categories, as would be possible with a larger sample. Further, the data we collected came when students were taking the IAT and not when they were in everyday situations where bias might be triggered. The Covid-19 pandemic interrupted the planned data collection and the next steps for research are to both collect data from more participants and to collect data when participants are engaged in real or simulated environments that might trigger bias to confirm the real-world applications of the work.

## 6   Conclusion

The purpose of this paper was to understand the relationship between bias as measured by the IAT and involuntary physiological responses. Our preliminary findings with a sample of undergraduate students both build on and expand the literature on the relationship between bias and physiological responses and the use of the wearable devices for understanding psychological constructs. These preliminary findings on the relationship between bias as measured by the IAT and physiological indicators have profound implications for bias mitigation and measurement and warrant further investigation.